\documentclass[lettersize,journal]{IEEEtran}
\usepackage{amsmath,amsfonts}
\usepackage{algorithmic}
\usepackage{array}
\usepackage[caption=false,font=normalsize,labelfont=sf,textfont=sf]{subfig}
\usepackage{textcomp}
\usepackage{stfloats}
\usepackage{url}
\usepackage{verbatim}
\usepackage{graphicx}

\usepackage{multirow}
\usepackage{booktabs}
\usepackage{enumitem}
\usepackage{color}
\usepackage{subfig}
\usepackage{float}
\usepackage{wrapfig}
\usepackage{tabularx}
\usepackage{mathbbol}
\usepackage{amsmath}
\usepackage{amssymb}
\usepackage{hyperref}
\usepackage[table]{xcolor}
\usepackage{colortbl}
\graphicspath{{figures/}}

\def\BibTeX{{\rm B\kern-.05em{\sc i\kern-.025em b}\kern-.08em
    T\kern-.1667em\lower.7ex\hbox{E}\kern-.125emX}}
\usepackage{balance}
\begin{document}
\title{Tackling Ambiguity from Perspective of Uncertainty Inference and Affinity Diversification for Weakly Supervised Semantic Segmentation}
\author{Zhiwei Yang, Yucong Meng, Kexue Fu, Shuo Wang, Zhijian Song
\thanks{Zhiwei Yang and Zhijian Song are with Academy for Engineering \& Technology, Fudan University, Shanghai 200433, China. Zhiwei Yang, Yucong Meng, Shuo Wang, and Zhijian Song are with Digital Medical Research Center, School of Basic Medical Science, Fudan University, and also with Shanghai Key Lab of Medical Image Computing and Computer Assisted
Intervention, Shanghai 200032, China. Kexue Fu is with Shandong Computer Science Center, China. Shuo Wang and Zhijian Song are the corresponding authors. (email: shuowang@fudan.edu.cn, zjsong@fudan.edu.cn).

This work is supported by National Natural Science Foundation of China (82072021), Shanghai Sailing Program (22YF1409300), CCF-Baidu Open Fund (CCF-BAIDU 202316) and International Science and Technology Cooperation Program under the 2023 Shanghai Action Plan for Science (23410710400).}}

\markboth{Journal of \LaTeX\ Class Files,~Vol.~18, No.~9, April~2024}%
{How to Use the IEEEtran \LaTeX \ Templates}

\maketitle

\begin{abstract}
Weakly supervised semantic segmentation (WSSS) with image-level labels intends to achieve dense tasks without laborious annotations. However, due to the ambiguous contexts and fuzzy regions, the performance of WSSS, especially the stages of generating Class Activation Maps (CAMs) and refining pseudo masks, widely suffers from ambiguity while being barely noticed by previous literature. In this work, we propose UniA, a unified single-staged WSSS framework, to efficiently tackle this issue from the perspective of uncertainty inference and affinity diversification, respectively. When activating class objects, we argue that the false activation stems from the bias to the ambiguous regions during the feature extraction. Therefore, we design a more robust feature representation with a probabilistic Gaussian distribution and introduce the uncertainty estimation to avoid the bias. A distribution loss is particularly proposed to supervise the process, which effectively captures the ambiguity and models the complex dependencies among features. When refining pseudo labels, we observe that the affinity from the prevailing refinement methods intends to be similar among ambiguities. To this end, an affinity diversification module is proposed to promote diversity among semantics. A mutual complementing refinement is proposed to initially rectify the ambiguous affinity with multiple inferred pseudo labels. More importantly, a contrastive affinity loss is further designed to diversify the relations among unrelated semantics, which reliably propagates the diversity into the whole feature representations and helps generate better pseudo masks. Extensive experiments are conducted on PASCAL VOC, MS COCO, and medical ACDC datasets, which validate the efficiency of UniA tackling ambiguity and the superiority over recent single-staged or even most multi-staged competitors. Codes will be publicly available \href{https://github.com/zwyang6/UniA.git}{here}.
\end{abstract}


\section{Introduction}

\IEEEPARstart{S}{emantic} segmentation, as a dense task of classifying every pixel of an input image, has enjoyed enormous popularity in many computer vision scenarios~\cite{1,2}.  Despite the remarkable success of semantic segmentation, fully-supervised semantic segmentation heavily relies on the numerous pixel-level annotated images, which are notoriously laborious and expensive to acquire. To alleviate the inherent reliance on labour-intensive pixel-level annotations, weakly- supervised semantic segmentation (WSSS) as an annotation-efficient alternative, has attracted increasing attentions. It aims to achieve dense segmentation tasks with more accessible labels, such as image-level labels~\cite{3,4,t125}, points~\cite{5}, scribbles~\cite{6,7}, and bounding boxes~\cite{8,9}. Among the above forms of annotations, image-level labels are the most accessible yet challenging, as they only provide cues on the existence of objects and offer the least semantic information.
\begin{figure}[t]
  \centering
  \includegraphics[width=8.7cm]{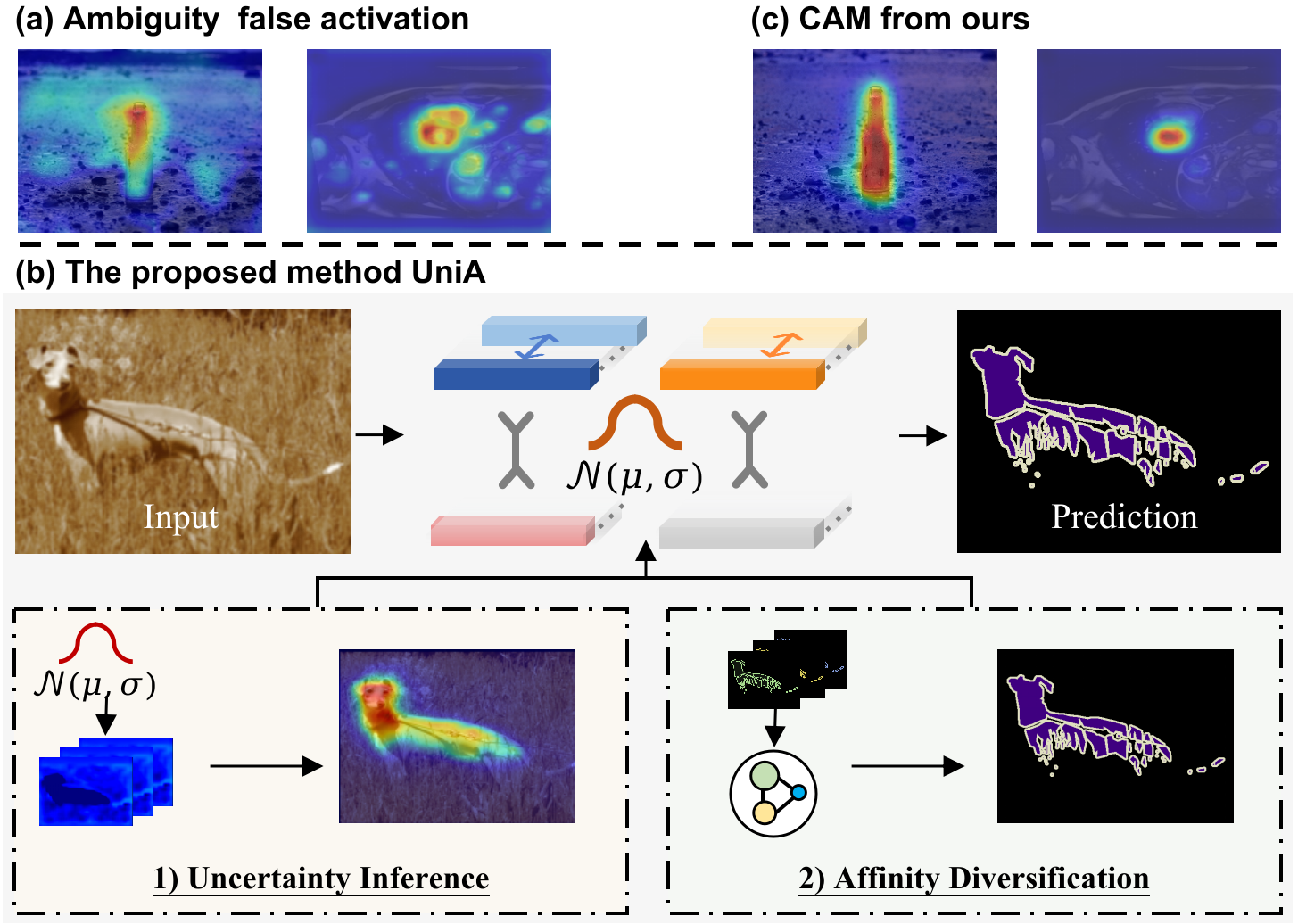}
   \caption{Motivation of UniA. (a) Ambiguity induces false estimation, which hinders the performance of WSSS. (b) The proposed UniA effectively suppresses it from the perspective of uncertainty inference and affinity diversification. (c) UniA can precisely activate objects.}
   \label{fig.1}
   \vspace{-1.5em}
\end{figure}

Commonly, most of the prevailing WSSS with image-level labels methods follow such a standard workflow: $1)$ generating CAMs~\cite{11} to localise objects by training a classification model; $2)$ taking CAMs as initial pseudo label seeds for further refinement~\cite{12,13}; $3)$ retraining a segmentation model with the supervision from pseudo labels and taking it as the final model to conduct inference~\cite{14}. In accordance with the pipeline, WSSS with image-level labels can be further grouped into multi-staged methods and the single-staged. Multi-staged WSSS methods~\cite{16,10,43,t124} intend to implement the workflow progressively. The classification and segmentation models need to be trained at different stages. Consequently, it holds better segmentation performance while being complicated and time-consuming during training. Single-staged methods~\cite{24,32,23,seco} train the classification and segmentation models at the same time by sharing the encoders of both models, thus the pipeline can be trained end-to-end. It is more efficient and easier to implement than the multi-staged while intends to have inferior segmentation performance. In this paper, we focus on the most challenging weakly-supervised semantic segmentation with image-level labels and train the pipeline end-to-end for efficiency.

Formally, the generation of CAMs and the refinement of pseudo masks determine the performance of both single-staged and multi-staged WSSS~\cite{t12,t123}. However, since the image-level labels only provide clues on the existence of objects, CAMs and the subsequent refinement mainly suffer from two critical issues: incompleteness and false estimation. The incompleteness issue stems from the fact that CAM only activates the most deterministic parts of objects and cannot fully exploit object contents. This issue has received many attentions and has been well studied from many novel perspectives~\cite{t1,13,50}. In opposition to the underestimation, false estimation is the issue that CAM intends to falsely activate the wrong regions distracted by the non-target objects from foregrounds and backgrounds. Most of the existing methods contribute this issue to the high co-occurring frequency of categories and tackle this issue from many intriguing insights, such as ~\cite{21} leveraging vision-language model CLIP~\cite{t2}, introducing out-of-distribution data~\cite{t3}, and data augmentation~\cite{43}, etc. 

However, we find that the false estimation issue does not only originate from the co-occurrence. The non-targets with a similar texture or color may still confuse the network and consequently result in the false estimation, as shown in Figure~\ref{fig.1} (a). The ambiguity between objects and background commonly exists in nature and medical images, which severely affects the quality of CAMs and pseudo masks, and hinders the performance of WSSS. However, this special type of false activation in WSSS is barely noticed. In this work, focusing on class activation and label refinement stages, we propose UniA, a unified single-staged WSSS framework, to efficiently tackle this issue from the perspective of uncertainty inference and affinity diversification, as shown in Figure~\ref{fig.1} (b). 

To reduce the ambiguity noise when generating CAMs, we argue that the ambiguity-induced false activation stems from the fact that the models cannot properly build the complex dependencies among fuzzy regions and are prone to overfit the unrelated objects. To this end, we introduce the uncertainty estimation to this stage for the first time and make the feature representation robust against noisy regions. Inspired by~\cite{u1,u2}, we model the feature extraction as a Gaussian distribution parameterised by two groups of probabilistic parameters. Specifically, we design two lightweight types of attention to represent the parameters from channel and spatial dimensions, which comprehensibly capture the semantic relations by locally perceiving the textures and globally building the dependencies among semantics. Then the variance from the distribution is estimated as uncertainty where high uncertainty represents predictions that dramatically vary at different inferences and are applied to infer the ambiguous regions. We also propose a learnable soft ambiguity masking strategy. It reduces the bias to unrelated semantics by gently incorporating the uncertainty perception into the features. To guarantee the distribution is closely in line with the feature representation, we further design a distribution loss to supervise the process and maintain the uncertainty level to capture the dependencies.

To further reduce the ambiguity noise, we notice that the prevalent affinity-based techniques~\cite{47,4,39} in the refinement stage suffer from the ambiguity and intend to be smooth. Recently, some works extract affinity from attention maps in ViT~\cite{t4} and conduct refinement in WSSS~\cite{24,16}. However, since the over-smoothness issue of ViT~\cite{t9,t10} and the high similarity among ambiguous regions and objects, the affinity in ViT is incompetent to measure the pairwise diversity among features, which may propagate the noisy pixels to the regions with similar color or textures and consequently results in the unexpected semantic errors during the refinement, detailed in Section~\ref{sec3.4}. To this end, we design an affinity diversification module to make the category semantics distinctive from the unrelated regions. We first propose a mutual complementing refinement to rectify the ambiguous affinity while saving the most reliable semantics from multiple inferred pseudo labels. However, such refinement based on comparisons may be tricky. Therefore, we further design a contrastive affinity loss to stably supervise the above process and make the affinity diversification learnable. It helps to promote the difference among ambiguous representations and generate more reliable pseudo labels.

In summary, our main contributions are listed as follows:
\begin{itemize}
\item[$\bullet$] The ambiguity-induced false activation issue in both stages of CAMs and refinement is formally reported in WSSS. A single-staged framework, named UniA, is proposed to efficiently tackle this issue from the perspective of uncertainty estimation and affinity diversification. 

\item[$\bullet$] The uncertainty estimation is introduced when generating CAMs for the first time and the feature extraction is novely modeled as a Gaussian distribution to avoid the bias to the unrelated regions. A distribution loss is designed to supervise the probabilistic process, guaranteeing the perceiving of complex dependencies among features.

\item[$\bullet$] An affinity diversifying module is designed to promote the difference among ambiguous regions. A mutual complementing refinement is proposed to rectify the ambiguous affinity while saving the most certain semantics. A contrastive affinity loss is designed to stably propagate such diversity to the feature representation and tackle the semantic errors from ambiguity.

\item[$\bullet$] Extensive experiments on PASCAL VOC 2012, MS COCO 2014, and medical ACDC 2017 datasets are conducted, demonstrating the superiority of UniA tackling ambiguity, as shown in Figure~\ref{fig.1} (c).
\end{itemize}

\section{Related Work}
\begin{figure*}[h!]
  \centering
  \includegraphics[width=18.2cm]{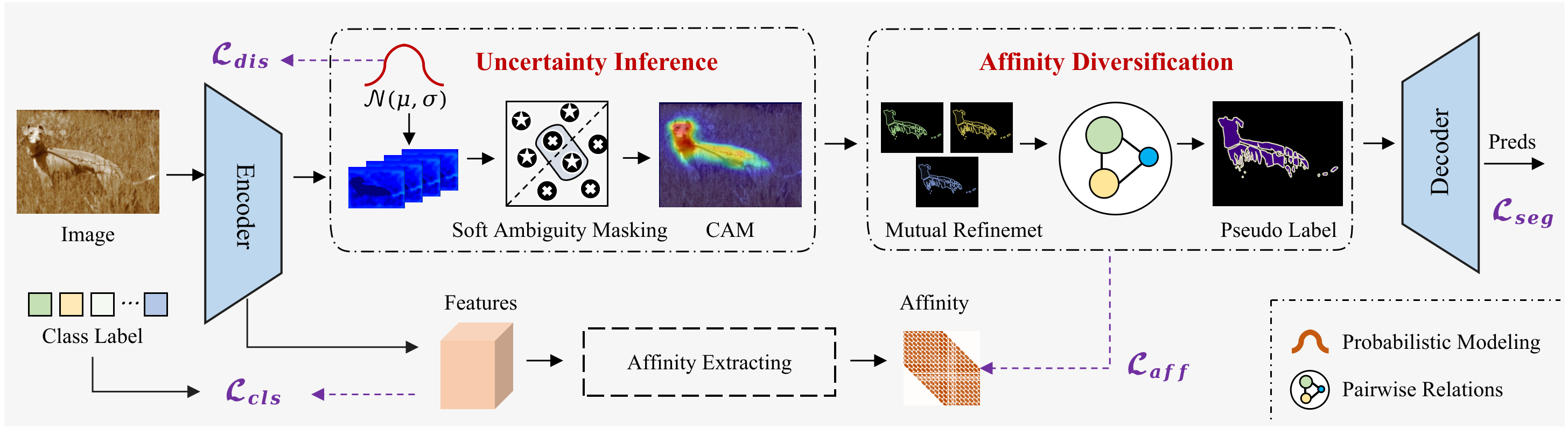}
   \caption{Overview of the proposed UniA for weakly supervised semantic segmentation. Given an input image, the uncertainty inference network firstly estimates uncertainty, which helps generate reliable CAM seeds. Then the affinity diversification module is designed to promote the difference among ambiguities, which further helps conduct reliable refinement. Finally, the pseudo labels are used to train a decoder for segmentation. The whole pipeline is trained end-to-end.}
   \label{fig.2}
  \vspace{-1em}
\end{figure*}
\subsection{Weakly-Supervised Semantic Segmentation}
CAM is a prevailing technique in WSSS with image-level labels. It commonly suffers two issues: incompleteness and false estimation. For the incompleteness issue, considerable efforts have been proposed from many intriguing perspectives, such as self-supervised learning~\cite{18,19}, contrastive learning ~\cite{20}, erasing~\cite{t1}, text-image matching~\cite{21,22}, and causal inference~\cite{48}. Recently, many transformer-based methods~\cite{16,24, t12} are proposed and achieve competent performance due to the inherent long-range dependency of self-attention. For the false estimation issue, most existing works attribute it to the high co-occurring frequency of objects and propose to tackle it by introducing external supervision~\cite{t2} or data~\cite{t3,43}. In this work, we observe that the ambiguity in both nature and medical images inevitably incurs false estimation as well and intend to tackle this issue from the perspective of uncertainty inference and affinity diversification.

\subsection{Uncertainty Learning in WSSS}
One concern of deep neural networks is that the model could occasionally output wrong predictions with high confidence~\cite{25}. One potential solution is uncertainty learning~\cite{26,27,u2}. Such paradigm is mainly featured with distribution modeling that each feature vector is a random variable, which is different from most studies that model the feature representation as a feature vector with fixed values. In fully supervised scenarios, ~\cite{u1} models the video processing as a Gaussian distribution and each frame is parameterized to be in line with the distribution. Inspired by it, \cite{33} model the image representation as a Gaussian distribution as well, and further estimate uncertainty to conduct robust detection and achieve success. In weakly supervised scenarios, some prior WSSS studies have leveraged uncertainty to estimate noise and achieved competitive performance.~\cite{28} learns the uncertainty within the bounding-box and performs online bootstrapping to generate segmentation masks. ~\cite{29} observes the noisy pixel in pseudo labels and applies uncertainty estimation to mitigate the noisy supervision signals. Unlike prior WSSS works which focus on modeling uncertainty during the segmentation stage, we introduce the uncertainty estimation in generating CAMs for the first time and highlight modeling the feature extraction as a probabilistic process when only image-level labels are available, which is more challenging while helpful to reduce the noise at the very beginning of WSSS. 

\subsection{Pairwise Modeling in WSSS}
Pairwise affinity modeling has been used to enhance the quality of pseudo labels in WSSS~\cite{47,4,39}. The classical CRF~\cite{12} leverages the pixel-level information to estimate the affinity relationship, which is commonly used as post-refinement in WSSS. However, due to the expensive pixel-level computing, it is not preferred in single-staged WSSS methods. Recently, some works~\cite{24,16} have noticed the consistency between affinity and attention maps in ViT and propose to extract pairwise relation from attention with minor computing overhead. However, due to the over-smoothness issue of ViT~\cite{t9,t10} and the similarity among ambiguous patches, the patch-wise affinity from attention suffers from the over-smoothness and holds similar semantic relations among ambiguity semantics and object representations. Consequently, the smooth affinity may propagate noisy pixels to ambiguous surroundings, leading to semantic error in refinement. In this work, complementary information is mined from a sequence of inferred pseudo labels to complement affinity from attention. A contrastive affinity loss is further designed to propagate the diversity into feature representations.

\section{Methodology}
In this section, the architecture of UniA is presented in Section \ref{sec3.1}. The CAM preliminary is introduced in Section \ref{sec3.11}. The proposed probabilistic modeling and uncertainty estimation are specifically introduced in Section \ref{sec3.2}. Finally, the affinity diversification module is detailed in Section \ref{sec3.4}. 
\subsection{Framework Overview}
\label{sec3.1}

Figure~\ref{fig.2} depicts a pipeline of UniA. Given $\mathcal{X}$ as the input image space and $\mathcal{Y}=\{1,2, \ldots, \mathrm{M}\}$ as the class label space, the training dataset is defined as ${D}=\left\{\left(x_{i}, y_{i}\right)\right\}_{i=1}^{n}$, where each tuple consists of an image $x_{i} \in \mathbb{R}^{3 \times \mathcal{H} \times \mathcal{W}}$ and the class label $y_{i} \in \mathcal{Y}$. For the input $x_{i}$, we adopt a transformer-based encoder to extract features, $x_{i} \rightarrow Z \in \mathbb{R}^{C \times H \times W}$, where $C$ is the feature channel, $H \times W$ is the spatial size. Then the feature $Z$ is modeled as a Gaussian distribution following~\cite{u1,u2}. Uncertainty is estimated from the variance of the distribution to infer the ambiguity regions and a soft ambiguity masking operation is conducted to suppress the false activation in a learnable manner. The whole estimation is guaranteed by a designed distribution loss $\mathcal{L}_{dis}$. To further reduce the ambiguity noise in the refinement stage, we propose an affinity diversity module to make the pairwise relations more diverse. A mutual complementing refinement is proposed to rectify the ambiguous affinity while saving the most certain semantics from multiple pseudo labels at first. A contrastive affinity loss $\mathcal{L}_{aff}$ is further designed to supervise the process and tackle the semantic errors from ambiguity. Finally, the generated pseudo labels are leveraged to train a segmentation network. Notably, encoders between classification and segmentation networks are shared so that the pipeline is trained end-to-end. 
\begin{figure*}
  \centering
  \includegraphics[width=18.2cm]{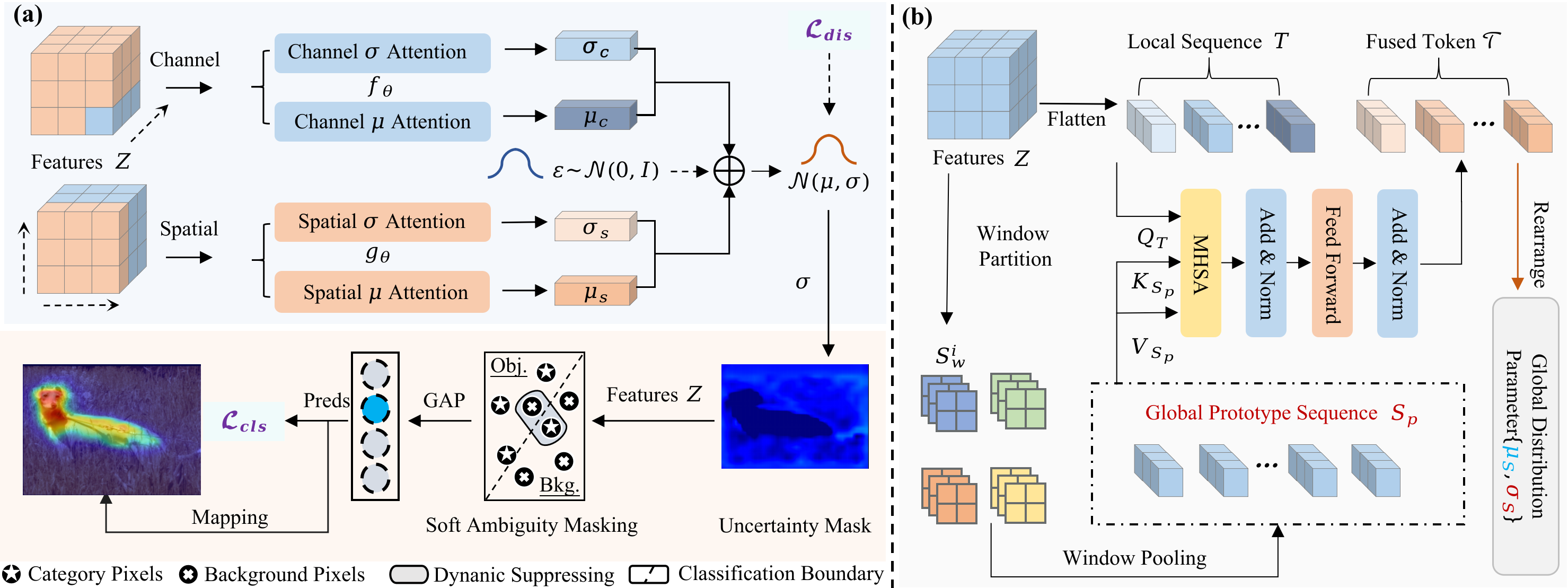}
   \caption{Architecture of uncertainty inference network. (a) Given the extracted features $Z$, our method learns a Gaussian distribution of $Z$ with channel and spatial attention, i.e., $f_{\theta}(\cdot)$ and $g_{\theta}(\cdot)$, for locally perceiving textures and globally building dependencies among semantics. Uncertainty is estimated from the distribution and a distribution loss $\mathcal{L}_{dis}$ is designed to guarantee it. The soft ambiguity masking is applied to incorporate the uncertainty into feature learning. Finally, the uncertainty-informed features are sent to a classification loss and CAM is generated. (b) Architecture of the spatial attention $g_{\theta}(\cdot)$.}
   \label{fig.3}
  \vspace{-1.0em}
\end{figure*}
\subsection{CAM Preliminary}
\label{sec3.11}
We first review the pipeline of generating CAMs. Given the input image $x_{i}$, a classification encoder is applied to extract features $Z$. Then the features are sent to a classification head before a global max pooling layer $\mathit{GMP}(\cdot)$. Then a multi-label soft margin loss is calculated to supervise the network:
\begin{equation}
    \begin{aligned}
    \mathcal{L}_{c l s}=
    \frac{1}{M}\sum_{j=1}^{M} 
    -y_{j} \log  \phi(z_{j})
    -(1-y_{j} ) \log (1- \phi(z_{j})),
    \end{aligned}
    \label{eq:9}
\end{equation}
where $ \phi(\cdot)$ is the sigmoid activation function, $z_{j}$ denotes the logits of $j$-th category, i.e., $z_{j}=\mathit{GMP}\left(Z\right)$. 
With the classifier weight $W \in \mathbb{R}^{C \times M}$ from the classification head, the CAM is generated after an activation function $Relu (\cdot)$: 
{\setlength\abovedisplayskip{5pt}
\setlength\belowdisplayskip{5pt}
\begin{equation}
    M_{CAM}=Relu\left(W^{T} Z\right),
    \label{eq:10}
\end{equation}}

\subsection{Uncertainty Inference}
\label{sec3.2}

 {\noindent\bf C.1) Probabilistic Distribution Modeling.} 

 We argue that the traditional models cannot properly build the complex dependencies among fuzzy regions and are prone to overfit the unrelated objects. Therefore, we introduce uncertainty learning to the feature representation and leverage it to suppress the false activation in CAMs. Specifically,  given the extracted features ${Z}$ from the encoder, we build the uncertainty inference network as a probabilistic representation. Following the popular setup in distribution learning~\cite{u1,u2,27}, the distribution is explicitly modeled as a Gaussian form, which means that ${Z}$ is regarded as a random variable and the value of each pixel is represented by a learnt distribution described by two groups of parameters, i.e., mean $\mu$ and variance $\sigma$. As shown in Figure~\ref{fig.3} (a), we send ${Z}$ to two parallel branches, i.e., mean encoder and variance encoder to model the parameters. To comprehensively model such distribution and capture the complex relations among features, we design two types of lightweight attention to represent the parameters from channel and spatial dimensions, which effectively captures the semantic relations by locally perceiving the textures and globally building the dependencies among semantics.

To locally extract the low-level details, such as texture or color, we leverage two $1\times1$ depth-wise convolutions, i.e., $f_\theta^\mu(\cdot)$ and $f_\theta^\sigma(\cdot)$, to build the distribution parameters $\mu_{c}$ and $\sigma_{c}$, respectively. The process can be formulated as:
\begin{equation}
    \mu_{c} = f_{\theta}^{\mu}(Z),
    \sigma_{c} = f_{\theta}^{\sigma}(Z),
    \label{eq:2}
\end{equation}
where $\mu_c $, $\sigma_c$ $\in \mathbb{R}^{C \times H \times W}$. Such local modeling effectively learns the fine-grained representation to perceive the difference among ambiguous regions.

To globally build the high-level semantic relations among ambiguities and objects, we design a lightweight window-based attention module for the distribution parameters $\mu_{s}$ and $\sigma_{s}$, as shown in Figure~\ref{fig.3} (b). Our motivation lies in the rich semantic correlation between local context and global context that helps to infer ambiguous regions. To this end, the features $Z$ are partitioned into windows $\left\{S_{w}^{i} \in\right. \left.\mathbb{R}^{C \times h \times w}\right\}_{i=1}^{N}$ at first, where $N$ is the number of windows, $h \times w$ is the size of each window. To capture the most representative features and filter the unrelated noise, we adopt window-max pooling to aggregate each window into a representative prototype: 
\begin{equation}
    S_{p}^{i}=\text { MaxPooling }\left(S_{w}^{i}\right).
    \label{eq:3}
\end{equation}
Then we gather all prototypes and get the prototype sequence $S_{p} \in \mathbb{R}^{C \times N}$ to include the global information. To conduct an efficient information interaction between local and global semantics, we then flatten the spatial $Z$ into a token sequence $T \in \mathbb{R}^{C \times (H \times W)}$, and leverage cross-attention to incorporate global semantic information into each local token from $T$. Such interaction considers the correlation of locality and global contexts and helps model complex relations among the ambiguities. The process is formulated as:
\begin{small}
\begin{equation}
    g_{\theta}\left(T, {S}_{p}\right)=\sum_{l=1}^{N} \frac{\exp \left(\left(Q_{T}\right)^{\top} K_{{S}_{p}}\right)}{\sqrt{C} \sum_{l=1}^{N} \exp \left(\left(Q_{T}\right)^{\top} K_{{S}_{p}}\right)} V_{{S}_{p}},
    \label{eq:4}
\end{equation}
\end{small}
where $g_{\theta}(\cdot)$ is the encoder to model distribution parameters $\mu_{s}$ and $\sigma_{s}$ along spatial dimension, respectively. $Q_{T}$ is the Query linearly projected from token $T$, $K_{{S}_{p}}$ and $V_{{S}_{p}}$ are the Key and Value linearly projected from ${S}_{p}$, respectively. Then the probabilistic parameters  are formulated as:
{\setlength\abovedisplayskip{9pt}
\setlength\belowdisplayskip{9pt}
\begin{equation}
    \mu_{s}=g_{\theta}^{\mu}\left(T, \boldsymbol{S}_{p}\right), \sigma_{s}=g_{\theta}^{\sigma}\left(T, \boldsymbol{S}_{p}\right),
    \label{eq:5}
\end{equation}}
where $\mu_{s}$, $\sigma_{s} \in \mathbb{R}^{C \times (H \times W)}$. Then we spatially rearrange the distribution parameters and get $\mu_{s}$, $\sigma_{s} \in \mathbb{R}^{C \times H \times W}$.

It should be noted that the learnt parameters are only feature vectors with fixed values at this time. To model a probabilistic distribution where each feature acts as a random variable, we integrate a standard Gaussian distribution $\varepsilon(\cdot)\sim \mathcal{N}(0, I)$ to make the whole process probabilistic and trainable~\cite{u2}. Then the distribution for features $Z$ is modeled:
\begin{equation}
    \mathcal{N}(\mu, \sigma)=\left(\mu_{c}+\mu_{s}\right)+\varepsilon\left(\sigma_{c}+\sigma_{s}\right),
    \label{eq:6}
\end{equation}
where $\mu=\mu_{c}+\mu_{s}, \sigma=\sigma_{c}+\sigma_{s}$.

 {\noindent\bf C.2) Uncertainty Quantification.} 
 
Inspired by~\cite{u3, 30, 33} that the predictions toward ambiguous regions intend to dramatically vary at different inferences, we directly leverage the probabilistic variance $\sigma$ to estimate uncertainty $U \in \mathbb{R}^{C \times H \times W}$ and apply it to suppress the ambiguous regions. 

However, with only $\mathcal{L}_{c l s}$ as the supervision, the variance $\sigma$ could unexpectedly decrease to zero and fail to perceive ambiguity. To this end, we further design a distribution loss with KL-Divergence to guarantee the uncertainty level and maintain the ambiguity inference. Following ~\cite{u1,u2}, we randomly draw $K$ maps $S_{k}\in\mathbb{R}^{K \times C \times H \times W}$ from the distribution $\mathcal{N}(\mu, \sigma)$ and leverage them as the predicted logits, instead of directly supervising the $\mu$ or $\sigma$. The reasons lie in two folds: 1) thanks to the incorporated standard distribution $\varepsilon(\cdot)$ in Equation~\ref{eq:6}, those samples could be viewed as a random variable continuously around the feature distribution, which is the essence of the probabilistic learning to avoid the overfitting issue~\cite{u1}. 2) Those samples are determined by $\mu$ and $\sigma$, which means that both probabilistic parameters can be supervised by the loss to guarantee the correct distribution. 

Considering the false activation by ambiguity, instead of directly leveraging the $M_{CAM}$ as the supervision, we propose to reweight CAMs by only saving the most certain semantics and filtering the uncertain with: 
{\setlength\abovedisplayskip{5pt}
\setlength\belowdisplayskip{5pt}
\begin{equation}
    M_{re}={\textstyle\sum_{i=1,j=1}^{W,H}} M_f(i,j)M_{CAM}(i,j),
    \label{eq:612}
\end{equation}
where $M_{f}=\mathbb{1}\left(M_{CAM}>\lambda\right)$ is the mask that only highlights strong-confident pixels in CAMs. $\mathbb{1}(\cdot)$ is an indicator function and $\lambda$ is a threshold value to filter the noisy pixels. After that, we generate a foreground-background pseudo mask $M_{bin}\in\mathbb{R}^{H \times W}$ from $M_{re}$ and take it as the supervision.

We average the samples along the channel to generate the probabilistic logit $S_{dis}\in\mathbb{R}^{H \times W}$ to present the output from the distribution. Finally, the distribution loss is formulated as:
{\setlength\abovedisplayskip{5pt}
\setlength\belowdisplayskip{5pt}
\begin{equation}
    \mathcal{L}_{dis}=KL(S_{dis}(\mu,\sigma)\left |  \right | M_{bin}),
    \label{eq:7}
\end{equation}}
where $\mathcal{L}_{dis}$ encourages the distribution to fit the certain features and reliably maintain the uncertainty inference. It should be noted that we also incorporate a regularization item $KL(\mathcal{N}(\mu, \sigma)\left |  \right | \mathcal{N}(0, I)) $ to regularize the distribution~\cite{u2}. 

{\noindent\bf C.3) Soft Ambiguity Masking.} 

With the estimated uncertainty, we incorporate it into the subsequent representation to further strengthen the ability to suppress the ambiguity. Specifically, we propose a learnable soft ambiguity masking to fuse the uncertainty. Firstly, we extract the certain information $Z_{\mathrm{c}}$ by $(1-U) \otimes Z $, where $\otimes$ denotes the Hadamard product. Then we incorporate the probabilistic uncertainty into features $Z$ by:
{\setlength\abovedisplayskip{5pt}
\setlength\belowdisplayskip{5pt}
\begin{equation}
    Z_{f}=\operatorname{softmax}\left(\frac{\left(Q_{Z}\right)^{\top} K_{Z_{c}}}{\sqrt{C}}\right) V_{Z_{c}},
    \label{eq:8}
\end{equation}}where $Q_{Z}$ is the Query linearly projected from the flattened $Z$, $K_{Z_{c}}$, $V_{Z_{c}}$ are the Key and Value from $Z_{\mathrm{c}}$, respectively. Such an incorporation process takes the uncertainty as learnable weights to dynamically exclude the irrelevant ambiguous features in a learnable manner. Finally, we leverage the uncertainty-guided feature $Z_{f}$ as the final feature and send it to the classification head and the segmentation head.

\vspace{-0.5em}
\subsection{Affinity Diversification}
\label{sec3.4}

{\noindent\bf D.1) Affinity Extraction.} 

 As illustrated in Figure~\ref{fig.5}, although the quality of pseudo labels is improved with raw affinity, there are still many failure cases where the over-smooth affinity propagates noisy pixels to the ambiguous surroundings and generates semantic false negatives and false positives. Therefore, we further propose an affinity diversification module to make the pairwise relations among semantics diverse. Firstly, let us review the process of affinity refinement in Vision Transformer~\cite{t4}. Due to the consistency between the attention map from Transformer and the pairwise relations in affinity, the attention map $M_{attn} \in \mathbb{R}^{HW \times HW}$ in multi-head self-attention (MHSA) is extracted as the affinity matrix. Notably, the affinity maps are symmetric. However, since the query and key are mapped into different spaces, the attention map in MHSA is asymmetric. To this end, we leverage Sinkhorn normalization~\cite{t11} to transform the map into a doubly stochastic matrix $M_{ds}$ and generate affinity maps $M_{aff}$ by averaging $M_{ds}$ and its transpose:

{\setlength\abovedisplayskip{5pt}
\setlength\belowdisplayskip{5pt}
\begin{equation}
    M_{aff} = (M_{ds} + M_{ds}^{T}) / {2}.
    \label{eq:11-2}
\end{equation}}

With affinity $M_{aff}$, random walk~\cite{7} is leveraged to refine CAM $M_{CAM}$ and generate the affinity-refined pseudo mask $P_{3}$. As shown in Figure~\ref{fig.5}, since the high similarity among ambiguities and objects, the smooth pairwise affinity negatively affects the pseudo masks at the same time. 
\begin{figure}[t]
  \centering
  \includegraphics[width=8.7cm]{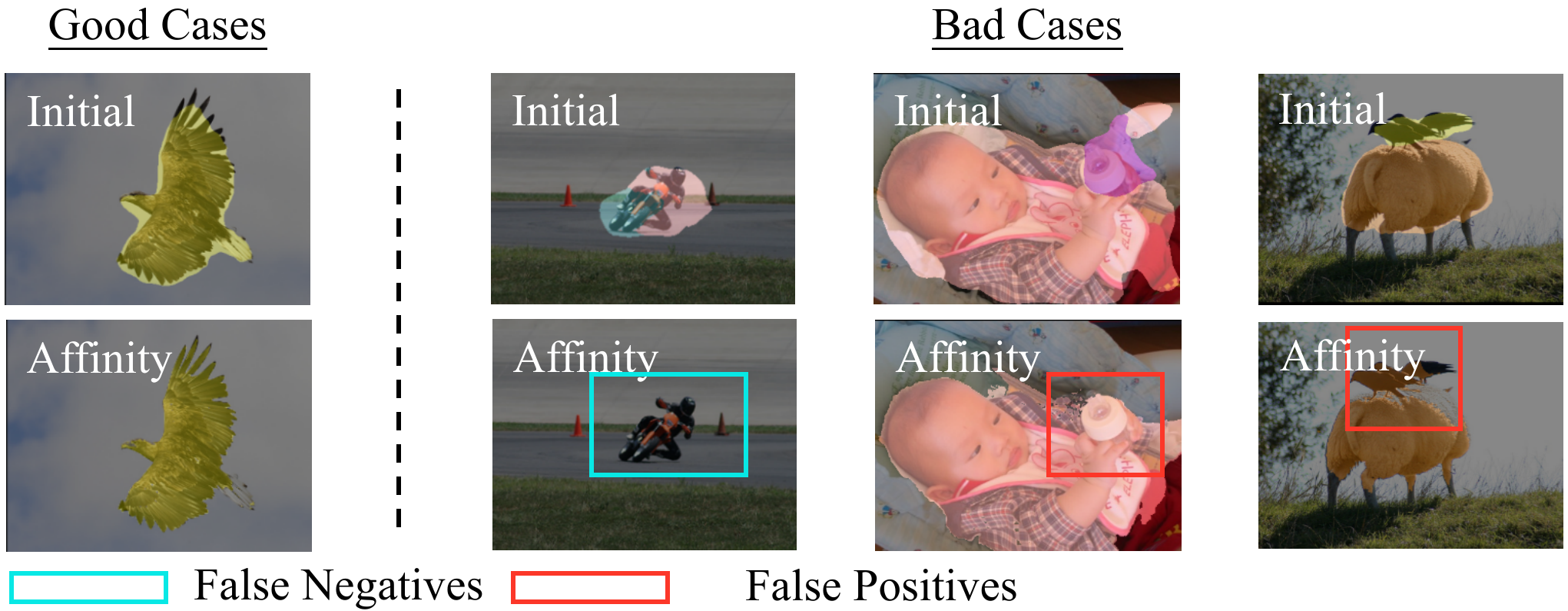}
   \caption{Pseudo labels refined with affinity. Although the raw affinity from attention improves the quality, it inevitably introduces noise and incurs false negatives and false positives due to ambiguity.}
   \label{fig.5}
  \vspace{-1.0em}
\end{figure}

\begin{figure}[t]
  \centering
  \includegraphics[width=8.7cm]{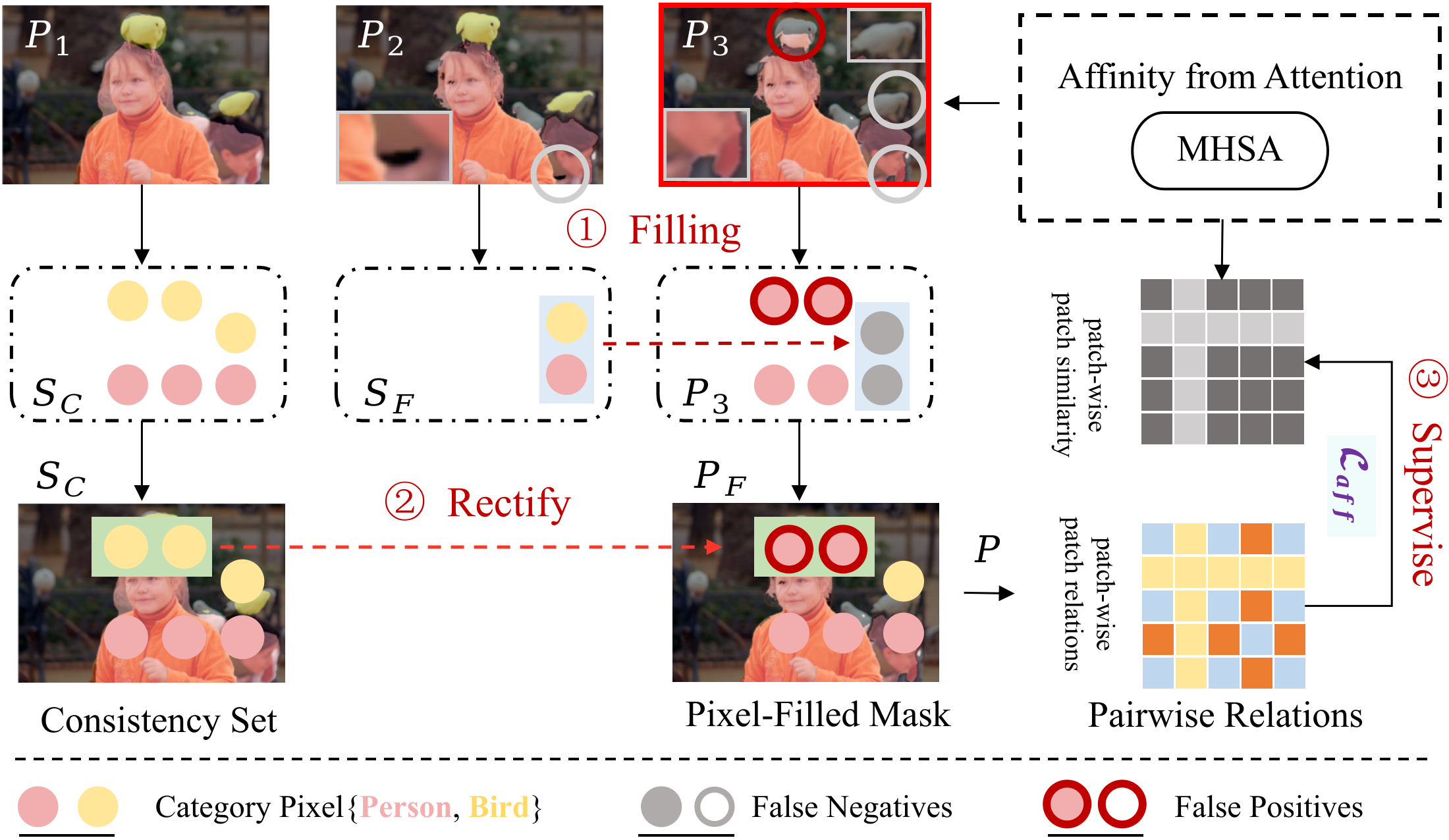}
   \caption{Illustration of Affinity diversification module. $P_{1}$ is the semantic mask generated from reweighted CAM. $P_{2}$ denotes the masks from RGB and position information. $P_{3}$ is the mask from affinity. Mutual complementing refinement directly remedies the pseudo labels and an affinity loss deeply propagates the diversity into feature representations .}
   \label{fig.4}
   \vspace{-1.0em}
\end{figure}
{\noindent\bf D.2) Mutual Complementing Refinement.} 

To this end, we first design a mutual complementing refinement to complement the raw affinity by mining diverse information from a sequence of pseudo masks, i.e., pseudo mask seeds $P_{1}$ directly from CAMs and $P_{2}$ from CAMs refined with RGB and spatial information by~\cite{24}, as shown in Figure~\ref{fig.4}. Our goal is to extract the diverse semantics from the pseudo label sequence to fix the semantic errors caused by ambiguity in $P_{3}$. It should be noted that since these two types of CAM-derived pseudo masks may still suffer from ambiguity, we further propose to reweight the initial CAM $M_{CAM}$ and only save the most reliable parts of $\{P_{1}, P_{2}\}$, as mentioned in Equation~\ref{eq:612}.

To tackle the false negatives issue in $P_{3}$, we extract the reliable true positives $S_{F}$ from $P_{2}$, as it is comparatively more complete, and apply $S_{F}$ to fix the false negatives in $P_{3}$:
{\setlength\abovedisplayskip{5pt}
\setlength\belowdisplayskip{5pt}
\begin{equation}
    S_{F}^{(i, j)}=\left\{\begin{array}{cc}
    P_{2}^{(i, j)}, & \text { if } P_{2}^{(i, j)} \in \mathcal{Y} \text {~\&~} P_{3}^{(i, j)} \in S_{B} \\
    0, & \text { otherwise, }
    \end{array}\right.
    \label{eq:11}
\end{equation}}

where $i, j$ is the spatial position in the maps. $\mathcal{Y}=\{1,2, \ldots, \mathrm{M}\}$ is the category information. $S_{B}=\{0,255\}$ is the prevalent setup when thresholding CAMs to pseudo masks~\cite{24, u4}, where $0,255$ denote the background and the ignored pixels, respectively. As illustrated in Figure~\ref{fig.4}, $S_{F}$ denotes the reliable true positive pixels from $P_{2}$ and is used to fill in the false negatives in $P_{3}$. Then we acquire the completed pseudo label $P_{F}$: 
\begin{equation}
    \vspace{-0.4em}
    P_{F}=M_{F} \otimes  P_{3}+S_{F},
    \label{eq:13}
\end{equation}
where the mask $M_{F}$ is used to remove the false negatives in $P_{3}$, which is denoted as $M_{F}^{(i, j)}= \mathbb{1}\left( S_{F}^{(i, j)} \notin \mathcal{Y}\right)$.

 To remedy the false positives issue, based on the locally correct property of $P_{1}$, we extract the consistency set $S_{C}$ from $P_{1}$ and $P_{2}$, which means that only the true positives in both $P_{1}$ and $P_{2}$ can be reliably used to rectify the false positives in $P_{3}$:
{\setlength\abovedisplayskip{2pt}
\setlength\belowdisplayskip{2pt}
\begin{equation}
    S_{C}^{(i, j)}= \left\{\begin{array}{cc}
    P_{1}^{(i, j)}, & \text { if } P_{1}^{(i, j)}=P_{2}^{(i, j)} \\ 
    0, & \text { otherwise },
    \end{array}\right.
    \label{eq:14}
\end{equation}
}
Finally, $S_{C}$ is leveraged to rectify the false positives and the more reliable pseudo labels are generated:
 {\setlength\abovedisplayskip{5pt}
\setlength\belowdisplayskip{5pt}
\begin{equation}
    P=M_{C} \otimes  P_{F}+S_{C},
    \label{eq:16}
\end{equation}}
where the mask $M_{C}$ is used to remove the false positives in $P_{3}$, which is denoted as $M_{C}^{(i, j)}= \mathbb{1}\left( S_{C}^{(i, j)} \notin \mathcal{Y}\right)$.

{\noindent\bf D.3) Affinity Diversifying Propagation.} 

However, simply rectifying the affinity masks based on the comparisons may still be tricky. To this end, we further design a contrastive affinity loss with remedied $P$ to supervise the affinity and make the whole process learnable, as presented in Figure~\ref{fig.4}. We intend to make the affinity more diverse by pulling together the affinity with the same semantics while pushing apart those among ambiguities. Specifically, if the pixel $(i,j)$ in $P$ has the same class label with pixel $(k,l)$, we view the corresponding affinity pairs in $M_{aff}$ as positive pairs; otherwise, the affinity pair is regarded as the negative. It forces the network to promote the difference between different objects and makes the attentions of categories more instinctive. Then the contrastive affinity loss $L_{aff}$ is designed to supervise the above process:
\begin{equation}
    L_{aff} = -\frac{1}{N^{+}}\sum_{q\in\mathcal{S^{+}}}\frac{e^{q/\tau }}{e^{q/\tau }+\sum_{(k)\in\mathcal{S^{-}}}e^{k/\tau }},
    \label{eq:16}
\end{equation}
where $N^{+}$ counts the positive pairs. $\mathcal{S^{+}}$ and $\mathcal{S^{-}}$ represent the set of positive and negative samples in $P$, respectively. $q$ and $k$ is the positive affinity logits and the negatives in $M_{aff}$. $\tau$ is the temperature factor to control the sharpness of contrast.

It should be noted that the effectiveness of the proposed affinity diversification module is two-fold. 1) Directly, it intends to mine complementary information from the mask sequence to comprehensively generate more reliable pseudo labels. 2) On the other hand, the remedied pseudo labels as a guidance to make the pairwise affinity in ViT more diverse and promote the difference among different features, which further improves the ability of representations in our encoders.

\subsection{Training Objectives}
Our framework comprises a multi-label soft margin loss $\mathcal{L}_{c l s}$ for classification, a distribution loss $\mathcal{L}_{dis}$ for uncertainty inference, and a contrastive affinity loss $\mathcal{L}_{a f f}$ for affinity diversification. It is noted that since UniA is a single-staged framework, a cross-entropy loss $\mathcal{L}_{{seg}}$ for segmentation is included as well. The overall loss for our framework is formulated as:
\begin{equation}
    \mathcal{L}_{UniA}=\mathcal{L}_{c l s}+\alpha \mathcal{L}_{dis} +\beta \mathcal{L}_{a f f}+\gamma \mathcal{L}_{{seg}},
    \label{eq:1}
\end{equation}
where $\alpha, \beta, \gamma$ are the loss weight factors to balance the corresponding contributions. Following previous works~\cite{u5,27,u4}, we also incorporate the regularization losses to enforce the consistency of CAMs and predicted masks.
\begin{figure*}
  \centering
  \includegraphics[width=18.3cm]{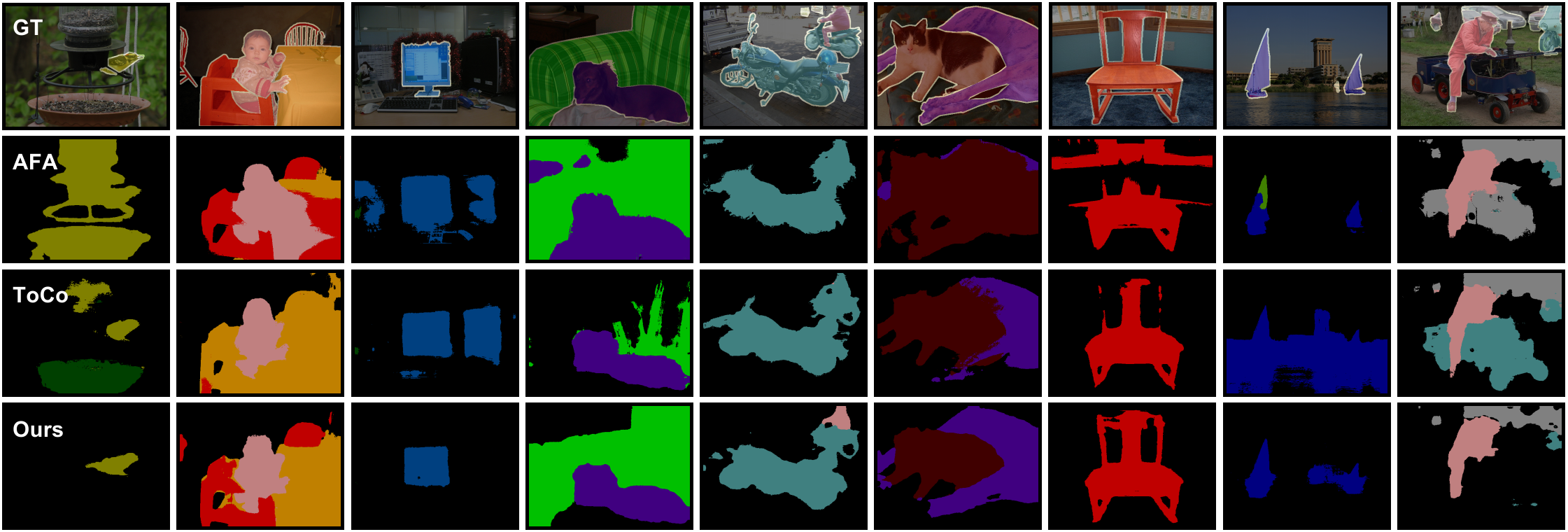}
   \caption{Qualitative segmentation results of UniA on PASCAL VOC 2012. AFA~\cite{24} and ToCo~\cite{u4} are adopted to draw a visual comparison with our method. UniA precisely segments objects and generates more complete predictions, especially for those objects with fuzzy boundaries.}
   \label{fig.7}
  \vspace{-1.0em}
\end{figure*}

\newcommand{\tablestyle}[2]{\setlength{\tabcolsep}{#1}\renewcommand{\arraystretch}{#2}\centering\footnotesize}
\begin{table}[!t]
\centering
\caption{Semantic segmentation comparison in terms of mIoU($\%$) on PASCAL VOC 2012 dataset. $Sup.$: supervision type. $\mathcal{F}$: full supervision. $\mathcal{I}$: image-level labels.$\mathcal{S}$: saliency maps.}
    \tablestyle{7.5pt}{1.1}
    \scalebox{1.}
    {
\begin{tabular}{lcccc}
\toprule
\multicolumn{1}{l|}{Method}                                & \multicolumn{1}{c|}{$Sup$.}                            & \multicolumn{1}{c|}{Backbone}                       & \multicolumn{1}{c|}{val}                                   & {test}                 \\ \midrule
\multicolumn{5}{l}{\textit{\textbf{Fully-supervised Models.}}} \\
\multicolumn{1}{l|}{DeepLabV2\cite{51}~{\scriptsize TPAMI'17}}              & \multicolumn{1}{c|}{}                & \multicolumn{1}{c|}{Res-101}                          & \multicolumn{1}{c|}{77.7}                             & -                 \\
\multicolumn{1}{l|}{WideResNet38\cite{38}~{\scriptsize PR'19}}        & \multicolumn{1}{c|}{$\mathcal{I}$}                  & \multicolumn{1}{c|}{WR-38}                          & \multicolumn{1}{c|}{80.8}                            & 82.5                 \\
\multicolumn{1}{l|}{Segformer\cite{37}~{\scriptsize NeurIPS'21}}              & \multicolumn{1}{c|}{}                & \multicolumn{1}{c|}{MiT-B1}                          & \multicolumn{1}{c|}{78.7}                           & -             
      \\ 
\multicolumn{1}{l|}{DeepLabV2\cite{51}~{\scriptsize TPAMI'17}}              & \multicolumn{1}{c|}{}                & \multicolumn{1}{c|}{ViT-B/16}                          & \multicolumn{1}{c|}{82.3}                             & -                 \\
      \midrule
\multicolumn{5}{l}{\textit{\textbf{Multi-stage WSSS Models.}}} \\
\multicolumn{1}{l|}{OAA+\cite{t20}~{\scriptsize ICCV'19}}               & \multicolumn{1}{c|}{}                & \multicolumn{1}{c|}{Res-101}                          & \multicolumn{1}{c|}{65.2}                                  & 66.4                 \\
\multicolumn{1}{l|}{ICD\cite{t21}~{\scriptsize CVPR'20}}               & \multicolumn{1}{c|}{}                & \multicolumn{1}{c|}{Res-101}                          & \multicolumn{1}{c|}{67.8}                                  & 68.0                 \\
\multicolumn{1}{l|}{AuxSegNet\cite{39}~{\scriptsize ICCV'21}}               & \multicolumn{1}{c|}{$\mathcal{I + S}$}                & \multicolumn{1}{c|}{WR-38}                          & \multicolumn{1}{c|}{69.0}                                  & 68.6                 \\
\multicolumn{1}{l|}{EPS\cite{40}~{\scriptsize CVPR'21}}                    & \multicolumn{1}{c|}{}                            & \multicolumn{1}{c|}{Res-101}                        & \multicolumn{1}{c|}{70.9}                                  & 70.8                 \\
\multicolumn{1}{l|}{L2G\cite{41}~{\scriptsize {CVPR'22}}}                    & \multicolumn{1}{l|}{}                                  & \multicolumn{1}{c|}{Res-101}                        & \multicolumn{1}{c|}{\textbf{72.1}}                         & \textbf{71.7}        \\ \midrule
\multicolumn{1}{l|}{SC-CAM\cite{42}~{\scriptsize CVPR'20}}                 & \multicolumn{1}{c|}{}                                  & \multicolumn{1}{c|}{WR-38}                          & \multicolumn{1}{c|}{66.1}                                  & 65.9                 \\
\multicolumn{1}{l|}{CDA\cite{43}~{\scriptsize ICCV'21}}                    & \multicolumn{1}{c|}{}                                  & \multicolumn{1}{c|}{WR-38}                          & \multicolumn{1}{c|}{66.1}                                  & 66.8                 \\
\multicolumn{1}{l|}{RIB\cite{10}~{\scriptsize NeurIPS'21}}                    & \multicolumn{1}{c|}{}                    & \multicolumn{1}{c|}{Res-101}                        & \multicolumn{1}{c|}{68.3}                                  & 68.6                 \\
\multicolumn{1}{l|}{ReCAM\cite{t13}~{\scriptsize CVPR'22}}               & \multicolumn{1}{c|}{{$\mathcal{I}$}}                                  & \multicolumn{1}{c|}{WR-38}                          & \multicolumn{1}{c|}{{68.5}}                         & {68.4}        \\
\multicolumn{1}{l|}{SIPE\cite{t14}~{\scriptsize CVPR'22}}                    & \multicolumn{1}{c|}{}                                  & \multicolumn{1}{c|}{Res-101}                        & \multicolumn{1}{c|}{68.8}                                  & 69.7                 \\ 
\multicolumn{1}{l|}{FPR\cite{t16}~{\scriptsize ICCV'23}}                    & \multicolumn{1}{c|}{}                                  & \multicolumn{1}{c|}{Res-101}                        & \multicolumn{1}{c|}{{70.3}}                                  & {70.1}            \\ 
\multicolumn{1}{l|}{MCTformer\cite{16}~{\scriptsize CVPR'22}}                    & \multicolumn{1}{c|}{}                                  & \multicolumn{1}{c|}{Res-101}                        & \multicolumn{1}{c|}{\textbf{71.9}}                                  & \textbf{71.6}                 \\ 
\midrule
\multicolumn{5}{l}{\textit{\textbf{Single-stage WSSS Models.}}}                                                                                                                                                                                   \\
\multicolumn{1}{l|}{1Stage\cite{44}~{\scriptsize CVPR'20}}                 & \multicolumn{1}{c|}{}                     & \multicolumn{1}{c|}{WR-38}                          & \multicolumn{1}{c|}{62.7}                                  & 64.3                 \\
\multicolumn{1}{l|}{AFA\cite{24}~{\scriptsize CVPR'22}}                      &  \multicolumn{1}{c|}{}                                & \multicolumn{1}{c|}{MiT-B1}                         & \multicolumn{1}{c|}{66.0}                                  & 66.3                 \\
\multicolumn{1}{l|}{WaveCAM\cite{t12}~\scriptsize TMM'23}                      & \multicolumn{1}{c|}{}                                  & \multicolumn{1}{c|}{MiT-B1}                         & \multicolumn{1}{c|}{66.2}                                  & 66.3                 \\
\multicolumn{1}{l|}{TSCD\cite{u7}~\scriptsize AAAI'23}                      & \multicolumn{1}{c|}{{$\mathcal{I}$}}                                  & \multicolumn{1}{c|}{MiT-B1}                         & \multicolumn{1}{c|}{67.3}                                  & 67.5                 \\
\multicolumn{1}{l|}{ViT-PCM\cite{t18}~\scriptsize ECCV'22}                      & \multicolumn{1}{c|}{}                                  & \multicolumn{1}{c|}{ViT-B/16}                         & \multicolumn{1}{c|}{70.3}                                  & 70.9                 \\
\multicolumn{1}{l|}{ToCo\cite{u4}~\scriptsize CVPR'23}                      & \multicolumn{1}{c|}{}                                  & \multicolumn{1}{c|}{ViT-B/16}                         & \multicolumn{1}{c|}{71.1}                                  & 72.2                 \\
\rowcolor[HTML]{EFEFEF} 
\multicolumn{1}{l|}{\cellcolor[HTML]{EFEFEF}\textbf{UniA (Ours)}} & \multicolumn{1}{c|}{\cellcolor[HTML]{EFEFEF}\textbf{}} & \multicolumn{1}{c|}{\cellcolor[HTML]{EFEFEF}ViT-B/16} & \multicolumn{1}{c|}{\cellcolor[HTML]{EFEFEF}\textbf{74.1}} & \textbf{73.6}        \\ \bottomrule
\end{tabular}
    }
    \label{table.1}
   \vspace{-1.5em}
\end{table}

\begin{table}[!t]
\centering
\caption{Evaluation and comparison of pseudo ground-truth in terms of mIoU ($\%$) on PASCAL VOC. $\dagger$: Our implementation. CRF~\cite{12} post-processing is applied in the evaluation.}
    \tablestyle{9pt}{1.15}
    \scalebox{1.}
    {
\begin{tabular}{lcllcllcll}
\toprule
\multicolumn{1}{l|}{Method}                                & \multicolumn{3}{c|}{Training images}                      & \multicolumn{3}{c}{train}                                 & \multicolumn{3}{c}{val}                                   \\ \midrule
\multicolumn{10}{l}{\textit{\textbf{Multi-stage WSSS Models.}}}                                                                                                                                                                              \\
\multicolumn{1}{l|}{CONTA \cite{48}~{\scriptsize NeurIPS'20}}                     & \multicolumn{3}{c|}{10K}                                   & \multicolumn{3}{c}{67.9}                                  & \multicolumn{3}{c}{-}                                      \\
\multicolumn{1}{l|}{AdvCAM \cite{50}~{\scriptsize CVPR'21}}                         & \multicolumn{3}{c|}{10K}                                  & \multicolumn{3}{c}{69.9}                                  & \multicolumn{3}{c}{-}                                     \\
\multicolumn{1}{l|}{MCTformer \cite{16}~{\scriptsize CVPR'22}}                      & \multicolumn{3}{c|}{10K}                                   & \multicolumn{3}{c}{69.1}                                  & \multicolumn{3}{c}{\textbf{-}}                            \\
\multicolumn{1}{l|}{CLIMS \cite{u8}~{\scriptsize CVPR'22}}                          & \multicolumn{3}{c|}{10K}                                   & \multicolumn{3}{c}{{70.5}}                         & \multicolumn{3}{c}{{-}}                            \\
\multicolumn{1}{l|}{W-OoD \cite{u9}~{\scriptsize CVPR'22}}                          & \multicolumn{3}{c|}{10K}                                   & \multicolumn{3}{c}{{72.1}}                         & \multicolumn{3}{c}{{-}}                            \\
\multicolumn{1}{l|}{AMN \cite{45}~{\scriptsize CVPR'22}}                          & \multicolumn{3}{c|}{10K}                                   & \multicolumn{3}{c}{{72.2}}                         & \multicolumn{3}{c}{{-}}                            \\
\multicolumn{1}{l|}{FPR \cite{t16}~{\scriptsize ICCV'23}}                          & \multicolumn{3}{c|}{10K}                                   & \multicolumn{3}{c}{{68.5}}                         & \multicolumn{3}{c}{{-}}                            \\
\multicolumn{1}{l|}{OCR \cite{14}~{\scriptsize CVPR'23}}                          & \multicolumn{3}{c|}{10K}                                   & \multicolumn{3}{c}{{69.1}}                         & \multicolumn{3}{c}{\textbf{-}}                            \\
\multicolumn{1}{l|}{CLIP-ES \cite{t19}~{\scriptsize CVPR'23}}                          & \multicolumn{3}{c|}{10K}                                   & \multicolumn{3}{c}{\textbf{75.0}}                         & \multicolumn{3}{c}{\textbf{-}}                            \\
\midrule
\multicolumn{10}{l}{\textit{\textbf{Single-stage WSSS Models.}}}                                                                                                                                                                             \\
\multicolumn{1}{l|}{1Stage \cite{49}~{\scriptsize CVPR'20}}                       & \multicolumn{3}{c|}{10K}                                   & \multicolumn{3}{c}{66.9}                                  & \multicolumn{3}{c}{65.3}                                  \\
\multicolumn{1}{l|}{AFA \cite{24}~{\scriptsize CVPR'2022}}                                               & \multicolumn{3}{c|}{10K}                                   & \multicolumn{3}{c}{68.7}                                  & \multicolumn{3}{c}{66.5}                                  \\
\multicolumn{1}{l|}{ViT-PCM\cite{t18}~{\scriptsize ECCV'22}}                                               & \multicolumn{3}{c|}{10K}                                   & \multicolumn{3}{c}{71.4}                                  & \multicolumn{3}{c}{-}                                  \\
\multicolumn{1}{l|}{ToCo\cite{u4}~{\scriptsize CVPR'23}}                                               & \multicolumn{3}{c|}{10K}                                   & \multicolumn{3}{c}{73.6}                                  & \multicolumn{3}{c}{72.3}                                  \\
\rowcolor[HTML]{EFEFEF} 
\multicolumn{1}{l|}{\cellcolor[HTML]{EFEFEF}\textbf{UniA (Ours)}} & \multicolumn{3}{c|}{\cellcolor[HTML]{EFEFEF}\textbf{10K}} & \multicolumn{3}{c}{\cellcolor[HTML]{EFEFEF}\textbf{75.9}} & \multicolumn{3}{c}{\cellcolor[HTML]{EFEFEF}\textbf{75.0}} \\ \bottomrule
\end{tabular}
    }
    \label{table.5}
   \vspace{-1.5em}
\end{table}
\section{Experiments}

\subsection{Experiment Setup}
{\noindent\bf Datasets and Evaluation Metrics.} The proposed approach is evaluated on three datasets, i.e., PASCAL VOC 2012\cite{34}, MS COCO 2014\cite{35}, and ACDC 2017\cite{36}. PASCAL VOC contains 21 semantic classes (including background). Following practice~\cite{1,24,44}, the dataset is augmented to $10,582$, $1,449$ and $1,456$ images for training, validating and testing, respectively. MS COCO 2014 dataset contains $81$ semantic classes,  $82,081$ images for the train set and $40,137$ images for val set. To verify the competence of UniA at objects with ambiguity, experiments are implemented on the medical ACDC 2017 dataset. It includes $100$ cases and is divided into $75$ cases for training and $25$ cases for testing. 

For evaluating metrics, mean Intersection-Over-Union (mIoU) is used as evaluation criteria for nature images. Dice Similarity Coefficient (DSC) is adopted for ACDC dataset. To further evaluate the competence at tackling false positives from ambiguity, confusion ratio (CR) is designed as the number of false positives / that of true positives, the lower and the better.  

{\noindent\bf Implementation Details.} 
Our encoder adopts ViT/B-16~\cite{t4} with a lightweight classification head and a LargeFOV segmentation head~\cite{u6}, which have presented a competitive performance on WSSS tasks. The parameters of the encoder are initialized with weights pre-trained on ImageNet\cite{52}. AdamW optimizer with an initial learning rate of $1\times6^{-5}$ is used and it decays with a polynomial scheduler. Furthermore, data augmentation strategies, such as scaling and random crop are adopted. For PASCAL VOC dataset, $20,000$ iterations are used for training. We extract $50$ samples from the learnt distribution to represent the distribution. For MS COCO dataset, $80,000$ iterations are adopted in total for training. The distribution modeling and affinity diversity module are implemented after $16,000$-th iteration. The weight factors $\alpha,\beta,\gamma$ are $0.1,0.1,0.15$, respectively. All the experiments in this work are implemented in RTX 3090.
\begin{figure*}
  \centering
  \includegraphics[width=18.3cm]{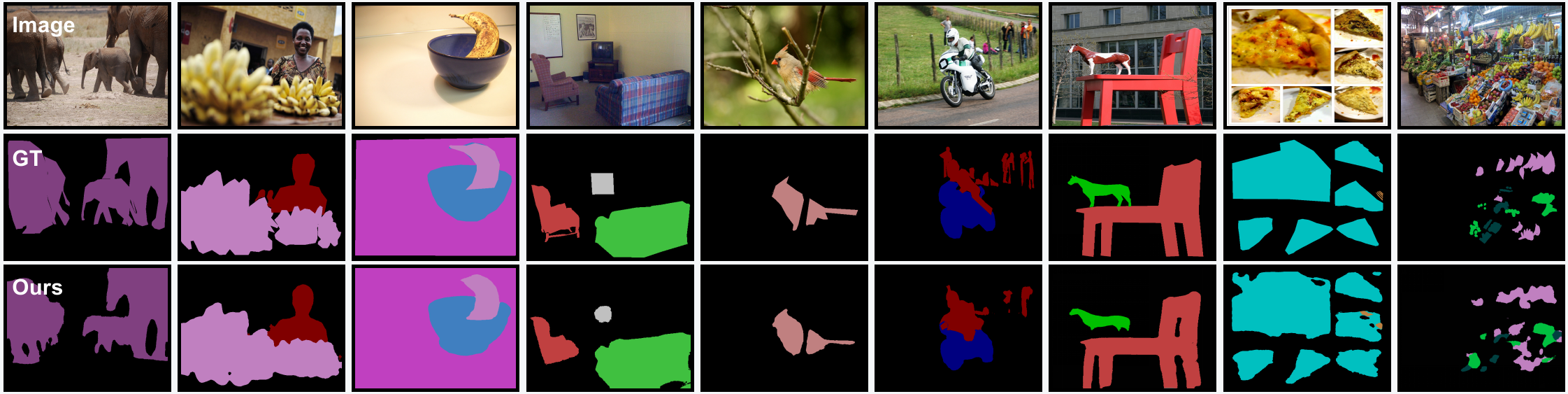}
   \caption{Qualitative segmentation results of UniA on MS COCO 2014 dataset. UniA is competent at generating high-quality semantic predictions.}
   \label{fig.6}
  \vspace{-0.6em}
\end{figure*}

\begin{table}[!t]
\centering
\caption{Semantic segmentation results on MS COCO dataset.}
    \tablestyle{10.5pt}{1.1}
    \scalebox{1.}{
    \begin{tabular}{lccc}
    \toprule
    \multicolumn{1}{l|}{Method}                                & \multicolumn{1}{c|}{\textit{Sup.}}             & \multicolumn{1}{c|}{Backbone}                       & val           \\ \midrule
    \multicolumn{4}{l}{\textit{\textbf{Multi-stage WSSS Models.}}}                                                                                                                   \\
    \multicolumn{1}{l|}{AuxSegNet\cite{39}~{\scriptsize ICCV'21}}               & \multicolumn{1}{c|}{}                          & \multicolumn{1}{c|}{WR-38}                          & 33.9          \\
    \multicolumn{1}{l|}{EPS\cite{40}~{\scriptsize CVPR'21}}                    & \multicolumn{1}{c|}{$\mathcal{I + S}$}                       & \multicolumn{1}{c|}{Res-101}                        & 35.7          \\
    \multicolumn{1}{l|}{L2G\cite{41}~{\scriptsize {CVPR'22}}}                    & \multicolumn{1}{l|}{}                          & \multicolumn{1}{c|}{Res-101}                        & \textbf{44.2} \\ \midrule
    \multicolumn{1}{l|}{SEAM\cite{19}~{\scriptsize CVPR'20}}                        & \multicolumn{1}{c|}{}                          & \multicolumn{1}{c|}{WR-38}                          & 31.9          \\
    \multicolumn{1}{l|}{CONTA\cite{48}~{\scriptsize NeurIPS'20}}                    & \multicolumn{1}{c|}{}                          & \multicolumn{1}{c|}{WR-38}                          & 32.8          \\
    \multicolumn{1}{l|}{CDA\cite{43}~{\scriptsize ICCV'21}}                    & \multicolumn{1}{c|}{}                          & \multicolumn{1}{c|}{WR-38}                          & 31.7          \\
    \multicolumn{1}{l|}{RIB\cite{10}~{\scriptsize NeurIPS'21}}                    & \multicolumn{1}{c|}{{$\mathcal{I}$}}       & \multicolumn{1}{c|}{Res-101}                        & 43.8          \\
    \multicolumn{1}{l|}{MCTformer\cite{16}~{\scriptsize CVPR'22}}               & \multicolumn{1}{c|}{}                          & \multicolumn{1}{c|}{WR-38}                         & {42.0} \\ 
    \multicolumn{1}{l|}{ESOL\cite{t15}~{\scriptsize NeurIPS'22}}               & \multicolumn{1}{c|}{}                          & \multicolumn{1}{c|}{WR-38}                         & {42.6} \\ 
    \multicolumn{1}{l|}{FPR\cite{t12}~{\scriptsize ICCV'23}}                    & \multicolumn{1}{c|}{}                          & \multicolumn{1}{c|}{Res-101}                        & \textbf{43.9}          \\ \midrule
    \textit{\textbf{Single-stage WSSS Models.}}               & \multicolumn{1}{l}{}                           &                                                     &               \\
    \multicolumn{1}{l|}{SLRNet\cite{t17}~{\scriptsize IJCV'22}}                      & \multicolumn{1}{c|}{}                         & \multicolumn{1}{c|}{WR-38}                         & 35.0          \\
    \multicolumn{1}{l|}{AFA\cite{24}~{\scriptsize CVPR'22}}                      & \multicolumn{1}{c|}{}                         & \multicolumn{1}{c|}{MiT-B1}                         & 38.9          \\
    \multicolumn{1}{l|}{WaveCAM\cite{t12}~{\scriptsize TMM'23}}                      & \multicolumn{1}{c|}{{$\mathcal{I}$}}                         & \multicolumn{1}{c|}{MiT-B1}                         & 39.5          \\
    \multicolumn{1}{l|}{TSCD\cite{u7}~{\scriptsize AAAI'23}}                      & \multicolumn{1}{c|}{}                         & \multicolumn{1}{c|}{MiT-B1}                         & 40.1          \\
    \multicolumn{1}{l|}{ToCo\cite{u4}~{\scriptsize CVPR'23}}                      & \multicolumn{1}{c|}{}                         & \multicolumn{1}{c|}{ViT-B/16}                         & 41.3          \\ 
    \rowcolor[HTML]{EFEFEF} 
    \multicolumn{1}{l|}{\cellcolor[HTML]{EFEFEF}\textbf{UniA (Ours)}} & \multicolumn{1}{c|}{\cellcolor[HTML]{EFEFEF}{}} & \multicolumn{1}{c|}{\cellcolor[HTML]{EFEFEF}ViT-B/16} & \textbf{43.2} \\ \bottomrule
    \end{tabular}
    }
    \label{table.40}
   \vspace{-1.5em}
\end{table}


\subsection{Main Results}
{\noindent\bf PASCAL VOC.}~Table~\ref{table.1} and ~Table~\ref{table.5} quantitatively present the segmentation performance and pseudo label quality compared to other impressive works on PASCAL VOC 2012 dataset, respectively. For the segmentation performance, the recent single-staged state-of-the-art (SoTA) works, ToCo~\cite{u4} generates $71.1\%$ and ViT-PCM~\cite{t18} achieves $70.3\%$ mIoU on VOC val set. Our framework achieves $74.1\%$ on val set. Notably, UniA can even achieve better results than multi-staged SoTA~\cite{16}. For a fair comparison, as reported in Table~\ref{table.1}, UniA achieves $90.0\%$ of its fully supervised counterpart ($82.3\%$ mIoU on PASCAL VOC val set)~\cite{51}, while ToCo achieves $86.4\%$ and multi-staged MCTFormer~\cite{16} achieves $89.0\%$ to its fully-supervised counterparts~\cite{51,38}, respectively. For the quality of pseudo labels, UniA achieves $75.9\%$ and $75.0\%$ mIoU on VOC train and val set, respectively, which significantly outperforms other single-staged competitors and is even superior to the multi-staged counterparts.

Moreover, the qualitative segmentation comparisons to the recent SoTA ToCo~\cite{u4} and AFA~\cite{24} are shown in Figure~\ref{fig.7}. It illustrates that since the proposed uncertainty inference and affinity diversification, the proposed method can fully localise the objects and avoid false positives. Both results show that UniA holds superiority over other recent competitors.

{\noindent\bf MS COCO.}~The segmentation performance of our UniA on MS COCO 2014 val set is presented in Table~\ref{table.40}. Our model achieves  $43.2\%$ mIoU on the val set, while the recent single-staged SOTAs, such as AFA~\cite{24} and ToCo~\cite{u4}, only achieves $38.9\%$ and $41.3\%$, respectively. 
For multi-staged competitors, UniA can achieve comparable or even better results. Visual results are presented in Figure~\ref{fig.6}. It demonstrates that UniA generates high-quality pseudo labels close to the ground truth.
\subsection{Ablation Studies and Analysis}
\begin{figure*}
  \centering
  \includegraphics[width=18.3cm]{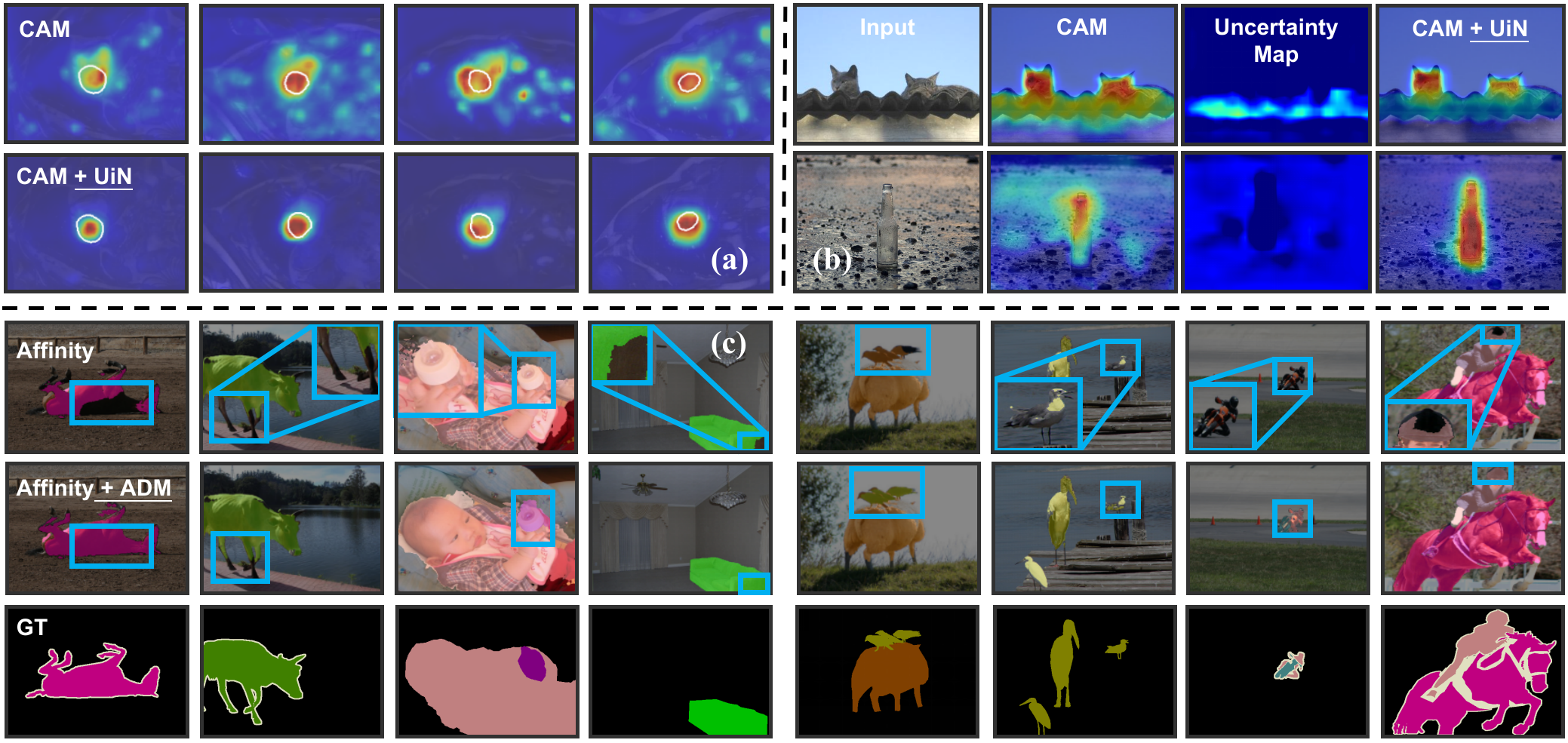}
   \caption{Ablative visualization of UniA. (a) Qualitative ablation of inference network (UiN). (b) Visualized uncertainty. (c) Qualitative ablation of the affinity diversification module (ADM). UiN suppresses the massive false activation and ADM rectifies the semantic errors (in the blue boxes) caused by ambiguity.}
   \label{fig.9}
  \vspace{-.5em}
\end{figure*}

\begin{table}[!t]
\centering
    \caption{Ablation study of UniA on VOC val set. UiN: uncertainty inference network. $\mathcal{L}_{dis}$: distribution loss. MCR: mutual complementing refinement. $\mathcal{L}_{aff}$: affinity diversity loss.}
    \tablestyle{4.8pt}{1.}
    \scalebox{1.}
    {
   \begin{tabular}{lccccccc}
    \toprule
    Method                & UiN & $\mathcal{L}_{dis}$ & MCR & $\mathcal{L}_{aff}$ & Precision & Recall & mIoU \\ \midrule
    Baseline              &     &      &     &      &-           &-        & 54.2 \\ \midrule
    \multirow{4}{*}{Ours} &\pmb{$\checkmark$}     &      &     &      & 80.6          &81.2        & 68.1 \\
                          &\pmb{$\checkmark$}     &\pmb{$\checkmark$}      &     &      & 81.3      & 83.6   & 71.0 \\
                          &\pmb{$\checkmark$}     &\pmb{$\checkmark$}      &\pmb{$\checkmark$}     &      & 81.4      & 84.7   & 71.8 \\
                          &\pmb{$\checkmark$}     &\pmb{$\checkmark$}      &\pmb{$\checkmark$}     &\pmb{$\checkmark$}      & 83.2      & 85.9   & 74.1 \\ \bottomrule
    \end{tabular}
    }
    \label{table.4}
   \vspace{-.5em}
\end{table}


\begin{table}[!t]
\centering
    \caption{Ablation of uncertainty inference network on VOC val set.}
    \tablestyle{5pt}{1.}
    \scalebox{1.}
    {
    \begin{tabular}{l|cccc}
    \toprule
    Method                & Channel Attention & Spatial Attention & Sampling & val  \\ \midrule
    \multirow{4}{*}{Ours} & \pmb{$\checkmark$}                & \pmb{$\checkmark$}                 &         & 72.3 \\
                          & \pmb{$\checkmark$}                &                 &\pmb{$\checkmark$}          & 72.9 \\
                          &                 &\pmb{$\checkmark$}                 &\pmb{$\checkmark$}          & 73.3 \\
                          & \pmb{$\checkmark$}                & \pmb{$\checkmark$}                & \pmb{$\checkmark$}        & 74.1 \\ \bottomrule
    \end{tabular}
    }
    \label{table.400}
   \vspace{-.5em}
\end{table}


{\noindent\bf Efficacy of Key Components.} Table \ref{table.4} quantitatively demonstrates the effectiveness of our method. Our baseline model, i.e., ViT-B/16, achieves $54.2\%$ mIoU on the PASCAL VOC val set. With the proposed uncertainty inference network (UiN) supervised only by classification loss, UniA improves the mIoU to $68.1\%$. With designed distribution loss $\mathcal{L}_{dis}$, the performance increases to $71.0\%$. This is because the distribution loss successfully maintains the uncertainty level. Then the mutual complementing refinement further improves the mIoU to $71.8\%$. With the supervision from the affinity loss $\mathcal{L}_{aff}$, UniA finally achieves $74.1\%$. It explains that the loss promotes the difference among ambiguities and propagates the diversity into feature learning and thus gains the improvement.

More specifically, we further explore the effectiveness of key strategies in the proposed uncertainty inference network, as reported in Table~\ref{table.400}. Since samples from the learnt distribution represent both probabilistic parameters in the distribution, directly taking $\mu$ to represent the distribution while no supervision to $\sigma$ degrades the performance to $72.3\%$. Moreover, modeling the distribution from the spatial dimension comprehensively captures the high-level relations among ambiguities and channel attention effectively perceives the local textures. Only modeling from channel attention or spatial attention drops the performance to $72.9\%$, $73.3\%$, respectively, which shows the effectiveness of the proposed modules.

\begin{table}[!t]
\centering
    \caption{Effectiveness of UniA tackling false activation on representative categories from VOC val set. Confusion ratio and IoU (in the bracket) are selected as the metrics.}
    \tablestyle{9pt}{1.1}
    \scalebox{1.}
    {
\begin{tabular}{l|cccl}
\toprule
       & AFA        & ToCo       & \multicolumn{2}{c}{Ours}                       \\ \midrule
plane   & 0.12~(79.3) & 0.19~(80.6) & \cellcolor[HTML]{EFEFEF}\textbf{0.08} & (87.0) \\
bird & 0.14~(79.8) & 0.42~(68.4) & \cellcolor[HTML]{EFEFEF}\textbf{0.07} & (89.1) \\
sheep  & 0.09~(84.8) & 0.06~(88.0) & \cellcolor[HTML]{EFEFEF}\textbf{0.03} & (90.0) \\
boat   & 0.42~(64.6) & 1.11~(45.4) & \cellcolor[HTML]{EFEFEF}\textbf{0.35} & (67.9) \\
cow    & 0.07~(83.9) & 0.03~(84.7) & \cellcolor[HTML]{EFEFEF}\textbf{0.02} & (87.1) \\
dog    & 0.12~(80.2) & 0.08~(83.7) & \cellcolor[HTML]{EFEFEF}\textbf{0.05} & (84.6) \\
plant  & 0.39~(51.8) & 0.59~(56.5) & \cellcolor[HTML]{EFEFEF}\textbf{0.29} & (65.0) \\
sofa  & 0.57~(44.6) & 0.77~(43.8) & \cellcolor[HTML]{EFEFEF}\textbf{0.35} & (51.2) \\
train  & 0.63~(59.6) & 0.75~(60.4) & \cellcolor[HTML]{EFEFEF}\textbf{0.52} & (62.7) \\
horse & 0.13~(76.0) & 0.09~(83.4) & \cellcolor[HTML]{EFEFEF}\textbf{0.05} & (84.1) \\ \midrule

All Categories& 0.36~(66.0) & 0.32~(71.1) & \cellcolor[HTML]{EFEFEF}\textbf{0.25} & (74.1) \\\bottomrule
\end{tabular}
    }
    \label{table.30}
   \vspace{-1em}
\end{table}


{\noindent\bf Effectiveness of Tackling False Activation.} To demonstrate the efficacy of UniA tackling false activation issue, some representative ambiguous categories on VOC val set are selected, such as wild categories (plant, sheep, bird, ambiguous to the background environment), sofa and plane (ambiguous to other categories). Confusion ratio (FP/TP, the lower, the better) and per-class IoU (in the bracket) are adopted as the metrics to evaluate the competence at suppressing false positives. As reported in Table~\ref{table.30}, UniA achieves higher IoU and lower confusion ratio on all selected cases compared to AFA~\cite{24} and the recent SoTA ToCo~\cite{u4}. UniA holds a confusion ratio of $0.35,0.35, 0.07$ for boat, sofa and bird, which is significantly lower than ToCo by $76.0\%,42.0\%, 35.0\%$. Moreover, the qualitative ablation and the uncertainty estimated in UniA are also visualized in Figure~\ref{fig.9} (a-b). It demonstrates that the uncertainty effectively suppresses false activation and the superiority of UniA at generating high-quality CAMs.

We also implement it on medical datastet ACDC, which features ambiguity (low contrast and similar textures). Confusion ratio (CR) and DSC are selected as the metrics to evaluate. As reported in Table~\ref{table.6}, incapable of inferring the ambiguous region, ToCo~\cite{24} achieves $80.36\%$ DSC and $0.49$ confusion ratio. Compared to it, UniA is more competent at tackling ambiguous images. Our method achieves better DSC of $83.75\%$ and CR of $0.21$, which significantly outperforms ToCo by $28\%$ in CR and even surpasses counterparts specially designed for medical images~\cite{46}. 

More than that, we further visualize CAMs from UniA and villain CAM on medical images. In Figure~\ref{fig.9} (a), the white boundaries represent the GT. As can be seen, ambiguous regions in medical images intend to confuse villain CAM and result in massive false activation. Compared to it, UniA significantly suppresses the falsely activated regions and precisely activates the objects, which demonstrates the effectiveness of UniA in tackling false activation. 

The significant improvements in qualitative and quantitative results may be explained that the designed uncertainty-guided representations effectively avoid overfitting to unrelated regions and the affinity diversification module successfully promotes the difference among ambiguity and objects.

{\noindent\bf Effectiveness of Affinity Diversification.} We compare the pseudo labels refined with affinity only and pseudo labels with our affinity diversification in Figure~\ref{fig.9} (c). It shows that simply leveraging affinity from attention inevitably incurs semantic errors, while the proposed method can address such issues well and generate more complete and precise pseudo labels by making the representations among ambiguities different.

\begin{table}[!t]
\centering
    \caption{Segmentation results on medical ACDC dataset. $\mathcal{ME}$: Designed for medical scenario. $\mathcal{N}$: Designed in nature scenario .}
    \tablestyle{5pt}{1.2}
    \scalebox{1.}
    {
\begin{tabular}{lcllccc}
\toprule
\multicolumn{1}{l|}{Method}                                & \multicolumn{3}{c|}{\textit{Sce.}}                     & \multicolumn{1}{c|}{Backbone}                 &{DSC(\%)}~{$\uparrow$} & {CR (\textit{\%})~{$\downarrow$}}      \\ \midrule
\multicolumn{7}{l}{\textit{\textbf{Multi-stage WSSS Models.}}}                                                                                                                                                 \\
\multicolumn{1}{l|}{SizeLoss\cite{SizeLoss}~{\scriptsize MIA'19}}                      & \multicolumn{3}{c|}{}                         & \multicolumn{1}{c|}{ENet}                           & 80.95                        & 0.31          \\
\multicolumn{1}{l|}{ISSOC\cite{46}~{\scriptsize PM\&B'21}}                      & \multicolumn{3}{c|}{{\multirow{-2}{*}{$\mathcal{ME}$}}}                        & \multicolumn{1}{c|}{VGG16}                               & 81.65                        & 0.26         \\ \midrule
\multicolumn{1}{l|}{AffinityNet\cite{1}~{\scriptsize CVPR'18}}                  & \multicolumn{3}{c|}{}                         & \multicolumn{1}{c|}{WR-38}                          & 80.17                        & 0.42          \\
\multicolumn{1}{l|}{IRNet\cite{47}~{\scriptsize CVPR'19}}                         & \multicolumn{3}{c|}{}                        & \multicolumn{1}{c|}{Res-50}                         & 74.67                        & 0.93          \\
\multicolumn{1}{l|}{BES\cite{BES}~{\scriptsize ECCV'20}}                         & \multicolumn{3}{c|}{\multirow{-2}{*}{$\mathcal{N}$}}                         & \multicolumn{1}{c|}{Res-101}                        & 77.53                        & 0.69          \\
\multicolumn{1}{l|}{CONTA\cite{48}~{\scriptsize NeurIPS'20}}                     & \multicolumn{3}{c|}{}                         & \multicolumn{1}{c|}{WR-38}                          & \textbf{83.51}                        & \textbf{0.23}          \\ \midrule
\multicolumn{7}{l}{\textit{\textbf{Single-stage WSSS Models.}}}                                                                                                                                                \\
\multicolumn{1}{l|}{AFA\cite{24}~{\scriptsize CVPR'22}}                         & \multicolumn{3}{c|}{}                        & \multicolumn{1}{c|}{MiT-B1}                         & 77.69                             & 1.08              \\
\multicolumn{1}{l|}{ToCo\cite{u4}~{\scriptsize CVPR'23}}                         & \multicolumn{3}{c|}{$\mathcal{N}$}                        & \multicolumn{1}{c|}{ViT-B/16}                         & 80.36                             & 0.49              \\
\rowcolor[HTML]{EFEFEF} 
\multicolumn{1}{l|}{\cellcolor[HTML]{EFEFEF}\textbf{UniA(Ours)}} & \multicolumn{3}{c|}{\cellcolor[HTML]{EFEFEF}{\multirow{-2}{*}{}}}  & \multicolumn{1}{c|}{\cellcolor[HTML]{EFEFEF}ViT-B/16} & \textbf{83.75}               & \textbf{0.21} \\ \bottomrule
\end{tabular}
    }
    \label{table.6}
   \vspace{-.5em}
\end{table}


\begin{table}[!t]
  \centering
  \caption{Impact of hyper-parameters on VOC val set. M: pseudo mask. Seg.: semantic prediction.}
    \subfloat[The sample number.]
  {
    \centering
    \begin{minipage}[b]{0.5\linewidth}{
    \begin{center}
    \tablestyle{5pt}{1.05}
    \scalebox{1}{
    \begin{tabular}{l|cc}
    \toprule
    Num Samples & M    & Seg  \\ \midrule
    10          & 73.9     & 72.7     \\
    30          & 74.5     & 73.5      \\
    \rowcolor[HTML]{EFEFEF} 
    \textbf{50} & {75.0} & \textbf{74.1} \\
    70          & \textbf{75.4}      & 73.7     \\ \bottomrule
    \end{tabular}
    }
    \end{center}
    }
    \end{minipage}
   }
  \subfloat[The loss weights.]
  {
    \centering
    \begin{minipage}[b]{0.5\linewidth}{
    \begin{center}
    \tablestyle{5pt}{1.05}
    \scalebox{1}{
    \begin{tabular}{l|cc}
    \toprule
    Loss Weights & M             & Seg           \\ \midrule
    0.8          & 75.5           &   72.5            \\
    0.5          & 74.1              & 73.3              \\
    \rowcolor[HTML]{EFEFEF} 
    \textbf{0.1} & \textbf{75.0} & \textbf{74.1} \\
    0.3          & 74.9              &  73.4             \\ \bottomrule
    \end{tabular}
    }
    \end{center}
    }
    \end{minipage}
  }
  \label{table.9}%
    \vspace{-1.5em}
\end{table}%

\begin{table}[!t]
\centering
\caption{Efficiency analysis of UniA compared to other methods on PASCAL VOC val set. M: multi-stage WSSS methods. S: single-stage WSSS method.}
    \tablestyle{6pt}{1.2}
    \scalebox{1.}
    {
    \begin{tabular}{l|ccccc}
    \toprule
    M          & CAM       & Refine   & \multicolumn{1}{c|}{Decode}   & val  & test \\ \cline{1-1} \cline{5-6} 
    CLIMS\cite{u8}      & 101mins   & 332mins  & \multicolumn{1}{c|}{635mins}  & 70.4 & 70.0 \\ \midrule
    S          &           &          &                               &      &      \\ \midrule
    AFA\cite{27}         & \multicolumn{3}{c|}{578mins}                         & 66.0 & 66.3 \\
    ToCo\cite{u4}        & \multicolumn{3}{c|}{524mins}                         & 71.1 & 72.2 \\
    \rowcolor[HTML]{EFEFEF} 
    \textbf{UniA(Ours)} & \multicolumn{3}{c|}{\cellcolor[HTML]{EFEFEF}\textbf{502mins}} & \textbf{74.1} & \textbf{73.6} \\ \bottomrule
    \end{tabular}
    }
\label{table.10}
\vspace{-.5em}
\end{table}

{\noindent\bf Hyper-parameter Sensitivity Analysis.} We generate the distribution logits by averaging the drawn samples and design a distribution loss to supervise it, thus the number of samples directly affects the probabilistic modeling. We explore the impact of the sample number on PASCAL VOC val set with mIoU, as reported in Table~\ref{table.9} (a). It is reported that the more samples drawn, the performance intends to be more stable since few samples cannot represent the distribution well. Considering the balance between the performance and the computing cost, UniA is competent to achieve favorable performance with $K = 50$ samples from the distribution. 

In addition, we verify the robustness by changing the loss weight factors of two designed losses $\mathcal{L}_{dis}$ and $\mathcal{L}_{aff}$. As reported in Table~\ref{table.9} (b), the performance of UniA stays consistent when those hyper-parameters vary, which shows that UniA is a robust WSSS framework to tackle the issue.

{\noindent\bf Training Efficiency Analysis.} We use uncertainty to infer the false activation of CAM and boost the performance of WSSS. One concern is that the extra sampling to present the distribution may introduce heavy computing overhead. To check this point, we evaluate the training speed by comparing the training time with some recent single-staged methods~\cite{24,u4} and multi-staged method~\cite{u8}. All the experiments are conducted on RTX 3090. As reported in Table~\ref{table.10}, CLIMS needs to be implemented progressively and it takes $1,068$ mins to finish the pipeline and achieve $70.4\%$ mIoU on VOC val set. Compared to it, thanks to the lightweight attentions to model the distribution and the single-stage paradigm, the sampling operation does not introduce much computing burden. UniA achieves $74.1\%$ mIoU on VOC val set with only $502$ mins, which is also more efficient than AFA and ToCo.

\section{Conclusion}
\label{sec:conclusions}
In this work, the false estimation caused by ambiguity is observed in WSSS. We propose a single-stage framework UniA to tackle this issue from the perspective of uncertainty inference and affinity diversification. To this end, the feature extraction is modeled as a probabilistic process reparameterised by parameters learnt from two lightweight attentions. A distribution loss is designed to guarantee the uncertainty learning and makes the representation robust to the irrelevant regions, which effectively helps capture the ambiguity and models the complex dependencies among features. In addition, an affinity diversification module is further proposed to promote the difference between ambiguity and object semantics. It initially rectifies the semantic errors with the mutual complementing refinement. Then an affinity diversifying loss is proposed to further propagate the diversity to the whole feature learning. Extensive experiments and analysis demonstrate the efficiency of UniA in tackling the ambiguity issue. 

One limitation is that the noise from ambiguity is only reduced in the stages of generating CAMs and the refinement in this work, while the noise is not directly considered when retraining a segmentation network. In the future, we hope to further boost the performance of UniA by designing denoising techniques to tackle ambiguity in the segmentation stage.


\bibliographystyle{IEEEtran}
\bibliography{IEEEabrv,references}



\end{document}